\documentclass[11pt,a4paper]{article}
\usepackage[hyperref]{rocling2023}
\usepackage{times}
\usepackage{latexsym}
\usepackage{booktabs}
\usepackage{graphicx}
\usepackage{CJKutf8}

\usepackage{microtype}

\roclingfinalcopy

\title{The North System for Formosa Speech Recognition Challenge 2023}

\author{Li-Wei Chen \begin{CJK*}{UTF8}{bsmi}陳力瑋\end{CJK*} \\
North Co., Ltd., Taiwan \\\And
Kai-Chen Cheng \begin{CJK*}{UTF8}{bsmi}鄭楷蓁\end{CJK*} \\
North Co., Ltd., Taiwan \\\And
Hung-Shin Lee \begin{CJK*}{UTF8}{bsmi}李鴻欣\end{CJK*} \\
North Co., Ltd., Taiwan}
\begin{document}
\maketitle

\begin{abstract}
This report provides a concise overview of the proposed North system, which aims to achieve automatic word/syllable recognition for Taiwanese Hakka (Sixian). The report outlines three key components of the system: the acquisition, composition, and utilization of the training data; the architecture of the model; and the hardware specifications and operational statistics. The demonstration of the system has been made public\footnote{\url{https://shorturl.at/mzGL7}}.
\end{abstract}


\section{Introduction}

This document furnishes a succinct yet comprehensive overview of the proposed North System, a technologically advanced mechanism designed with the primary objective of achieving automatic recognition of words and syllables specific to the Taiwanese Hakka language, with a particular focus on the Sixian dialect. The report meticulously delineates three pivotal components integral to the effective functionality and operation of the system, as enumerated below:

\begin{enumerate}
\item Acquisition, Composition, and Utilization of Training Data:
\begin{itemize}
\item Acquisition: The systematic collection and sourcing of relevant linguistic data pertinent to the Taiwanese Hakka language, ensuring a robust and representative dataset.
\item Composition: The strategic assembly and organization of the acquired data, ensuring it is structured in a manner conducive to effective machine learning.
\item Utilization: The application of the composed data in training the system, ensuring it accurately and efficiently recognizes and processes linguistic elements of the Sixian dialect.
\end{itemize}

\item Architectural Framework of the Model:
\begin{itemize}
\item A detailed exposition of the structural and operational framework of the model, elucidating the technological and algorithmic methodologies employed to facilitate accurate linguistic recognition and processing.
\item An exploration of the model’s design principles, underlying algorithms, and computational processes that enable it to effectively learn, recognize, and interpret the linguistic nuances of the Taiwanese Hakka language.
\end{itemize}

\item Hardware Specifications and Operational Metrics:
\begin{itemize}
\item Hardware Specifications: A thorough breakdown of the technological infrastructure supporting the system, detailing the hardware components and specifications that underpin its operation.
\item Operational Metrics: An analytical overview of the system’s performance metrics, providing insights into its operational efficiency, accuracy, and reliability in real-world applications.
\end{itemize}
\end{enumerate}

\section{Data}

Table 1 meticulously presents a detailed inventory of our principal corpus sources, accompanied by their respective statistical details, providing an in-depth insight into the voluminous data utilized in our research and development endeavors.

In addition to the primary corpus, our research team has assiduously gathered an extensive collection of Hakka (Sixian dialect)-related speech data, exceeding 100 hours, from a variety of online platforms. This includes, but is not limited to:

\begin{itemize}
\item YouTube: A prominent video-sharing platform where a myriad of Hakka (Sixian) linguistic content, ranging from casual conversations to formal discourses, has been extracted.
\item Podcasts: Audio programs and series that provide a rich source of conversational and narrative Hakka (Sixian) speech data.
\item Additional Online Platforms: Various other digital platforms that host a wealth of linguistic content pertinent to the Hakka (Sixian) dialect.
\end{itemize}

Moreover, we have amassed a substantial volume of Hakka text data, specifically curated for language modeling purposes, from a plethora of websites ardently dedicated to the promotion, preservation, and dissemination of the Hakka language and its diverse dialects.

Our methodologies for obtaining speech data are not solely confined to direct data gathering but also partially derive inspiration from the scholarly paper presented at O-COCOSDA 2020 by Dr. Hung-Shin Lee \cite{lee2020}. This paper provides valuable insights and methodologies that have been judiciously considered and adapted to enhance our data acquisition strategies.

It is imperative to underscore that our research and development team consciously opted to abstain from employing speech synthesis as a means for generating training speech data. This decision is firmly rooted in our belief that, while speech synthesis may offer a modicum of utility in certain contexts, its efficacy is notably constrained in the realm of speech recognition due to the intrinsic emphasis on accommodating and understanding variability in speaker characteristics and environmental acoustics.

\begin{table}[ht]
\caption{Statistics of sources of Hakka (Sixian) speech. The speech channels used in Official (train) and Official (pilot-test) are lavalier and Zoom, respectively. Hakka Dictionary comes from Dictionary of Frequently-Used Taiwan Hakka (\url{https://hakkadict.moe.edu.tw}). HAC, provided by Hakka Affairs Council (\url{https://corpus.hakka.gov.tw}), is further cleaned by our technology. SPU denotes average seconds per utterance.}
\vspace{-5pt}
\label{tab:one_class}
\small
\centering
\scalebox{1.0}{
\begin{tabular}{lccc}
\toprule
\bf Source & \bf Hours & \bf \# Utt. & \bf SPU \\
\toprule
Official Dataset (train) & 59.43 & 20,591 & 10.39 \\
Official Dataset (pilot-test) & 10.01 & 3,595 & 10.02 \\
Hakka Dictionary & 5.84 & 15,250 & 1.38 \\
HAC & 11.26 & 4,216 & 9.61 \\
\midrule
Total & 86.54 & 43,652 & 7.14 \\
\bottomrule
\end{tabular}}
\vspace{-10pt}
\end{table}

\section{Model Structure}

The meticulous training of the acoustic model necessitates the concatenation of two distinct types of speech features: the 40-dimensional Mel Frequency Cepstral Coefficients (MFCCs) and the 1024-dimensional Semi-Supervised Learning (SSL) embeddings. The SSL model, which was previously subjected to training on a comprehensive dataset of Chinese linguistic data utilizing the HuBERT-large model, is judiciously utilized to procure the SSL embeddings.

The model employed is fundamentally based on two primary architectural structures. Firstly, the Chain-based Discriminative Autoencoder (DcAE) \cite{lee2022}, and secondly, the Multistream Convolutional Neural Network (CNN) \cite{han2021}. For an in-depth elaboration and comprehensive understanding of these structures, readers are ardently encouraged to consult the referenced scholarly papers or the Appendix at the end of the paper. The integration of both aforementioned structures is employed in a joint training methodology, wherein the latter structure serves as the foundational bedrock upon which the former is developed and refined. The overarching objective permeating the entirety of the model is to assiduously minimize losses associated with Automatic Speech Recognition (ASR), particularly those pertaining to lattice-free Maximum Mutual Information (MMI), as well as errors inherent in feature reconstruction and restoration. Through a meticulous analysis of speech and noise factors, and the incorporation of multi-resolution information, the model significantly enhances its performance and robustness in various linguistic contexts.

The rescoring mechanism, which is meticulously designed to operate on the word lattice generated by a four-gram language model, employs a Recurrent Neural Network Language Model (RNN-LM) to enhance the accuracy and reliability of linguistic predictions and outputs.

It is pivotal to note that, consequent to the insufficiency of Graphical Processing Unit (GPU) resources, our team made a conscious decision to abstain from proceeding with the optimization and fine-tuning of model parameters and hyperparameters. This includes, but is not limited to, the number of model layers, training epochs, batch size, learning rate, among other crucial variables that significantly impact the model’s learning and predictive capabilities.

\section{Hardware}

In the execution of our experiment, we strategically employed a total of four NVIDIA RTX™ A5000 graphics cards, which are renowned for their robust computational capabilities and adeptness in handling graphically-intensive tasks and processes. It is imperative to underscore that this allocation of graphical processing units (GPUs) was the sole resource utilized to facilitate the extensive computational demands inherent in the training of the neural model.

The entirety of the training duration for a singular neural model was approximately 82 hours, a substantial temporal investment that underscores the computational and temporal demands associated with the development and refinement of sophisticated neural networks. This extensive training period was meticulously undertaken to ensure the model was afforded ample opportunity to learn, adapt, and refine its predictive capabilities, thereby enhancing its overall performance and reliability in practical applications.

\section{Final Results}

The conclusive results, which encapsulate the outcomes and findings derived from our experimental processes, are meticulously presented in Table 2. Our North system got the double champion of this Formosa Speech Recognition Challenge 2023 (General Group) in the two tracks of Hakka Character and Pinyin. It is imperative to underscore that one of the predominant weaknesses that permeated our model during the training phase is the conspicuous absence of spontaneous speech. This deficiency in the training data potentially impacts the model's capacity to accurately and reliably recognize and process unscripted, natural linguistic patterns and variations, thereby presenting an area warranting further investigation and enhancement in future research endeavors.

We also compared the Hakka ASR system developed by ASUS\footnote{\url{https://www.asuscloud.com/event-hakka}}, which is built up based on OpenAI's Whisper. Table \ref{tab:guide2} shows that our system is far superior to ASUS. It is worth noting that ASUS's system cannot handle sentences that are too long, meaning that 384 utterances could not be processed and were not included in the error rate calculation.

\begin{table}[t]
\caption{Final results of North ASRs with respect to two tracks, \begin{CJK*}{UTF8}{bsmi}客語漢字\end{CJK*} (Hakka Character) and \begin{CJK*}{UTF8}{bsmi}客語拼音\end{CJK*} (Hakka Pinyin), in terms of character and syllable error rate (\%), respectively. The proportion of reading and spontaneous speech in the total evaluation data is approximately 13\% and 87\% respectively. The total number of utterances for evaluation is 5,913.}
\vspace{0pt}
\label{tab:guide}
\small
\centering
\setlength{\tabcolsep}{9pt}
\begin{tabular}{lcccc}
\toprule
\textbf{Track} & \textbf{Read} & \textbf{Spont.} & \textbf{Average} & \textbf{Rank} \\ 
\midrule\midrule
\begin{CJK*}{UTF8}{bsmi}客語漢字\end{CJK*} & 4.27 & 19.14 & 17.15 & 1 \\
\begin{CJK*}{UTF8}{bsmi}客語拼音\end{CJK*} & 7.33 & 18.90 & 17.42 & 1 \\
\bottomrule
\end{tabular}
\end{table}

\begin{table}[t]
\caption{Comparisons of North ASRs and ASUS Hakka ASRs respect to two tracks, \begin{CJK*}{UTF8}{bsmi}客語漢字\end{CJK*} (Hakka Character) and \begin{CJK*}{UTF8}{bsmi}客語拼音\end{CJK*} (Hakka Pinyin), in terms of character and syllable error rate (\%), respectively. Because the ASUS Hakka ASR cannot deal with long utterances, only 5,529 utterances are evaluated.}
\vspace{0pt}
\label{tab:guide2}
\small
\centering
\setlength{\tabcolsep}{9pt}
\begin{tabular}{lccc}
\toprule
\textbf{Track} & \textbf{ASUS} & \textbf{North} & \textbf{Rel. Improve.} \\ 
\midrule\midrule
\begin{CJK*}{UTF8}{bsmi}客語漢字\end{CJK*} & 28.87 & \bf 18.17 & 37.06 \\
\begin{CJK*}{UTF8}{bsmi}客語拼音\end{CJK*} & 42.43 & \bf 19.65 & 53.69 \\
\bottomrule
\end{tabular}
\end{table}

\section*{Acknowledgments}
Regarding the collection and understanding of Hakka language corpus, we would like to express our gratitude to the following individuals for their professional assistance:
\begin{itemize}
\item \begin{CJK*}{UTF8}{bsmi}黃麗芬（北一女中客語教師）\end{CJK*}
\item \begin{CJK*}{UTF8}{bsmi}徐煥昇（育英國小龍騰分校教師）\end{CJK*}
\item \begin{CJK*}{UTF8}{bsmi}張陳基（國立聯合大學文化創意與數位行銷學系教授）\end{CJK*}
\end{itemize}

\section*{Appendix}
\subsection*{Multistream CNN \cite{han2021}}
The Multistream Convolutional Neural Network (CNN), a pioneering neural network architecture, meticulously designed to fortify acoustic modeling within the realm of speech recognition tasks, is introduced. The architectural framework processes input speech utilizing diverse temporal resolutions, achieved by applying varying dilation rates to convolutional neural networks across multiple streams, thereby attaining robustness in acoustic modeling. The dilation rates are judiciously selected from multiples of a sub-sampling rate, specifically, three frames. Each stream systematically stacks Time Delay Neural Network-F (TDNN-F) layers, a variant of 1D CNN, and the output embedding vectors derived from the streams are concatenated and subsequently projected to the terminal layer.

The efficacy of the Multistream CNN architecture is validated through demonstrable and consistent enhancements against Kaldi’s optimal TDNN-F model, observed across a myriad of data sets. The Multistream CNN facilitates a 12\% (relative) improvement in the Word Error Rate (WER) of the test-other set within the LibriSpeech corpus. Furthermore, on custom data derived from ASAPP’s production Automatic Speech Recognition (ASR) system for a contact center, it records a relative WER enhancement of 11\% for customer channel audio, thereby substantiating its robustness to data in uncontrolled environments. In terms of the real-time factor, the Multistream CNN surpasses the baseline TDNN-F by 15\%, thereby also indicating its practical applicability within production systems. When amalgamated with self-attentive Simple Recurrent Unit (SRU) Language Model (LM) rescoring, the Multistream CNN significantly contributes to ASAPP achieving an optimal WER of 1.75\% on the test-clean set in LibriSpeech.

\subsection*{Chain-based Discriminative Autoencoder \cite{lee2022}}
In preceding research endeavors, the authors introduced a model known as the Discriminative Autoencoder (DcAE), specifically tailored for applications within the domain of speech recognition. The DcAE amalgamates two distinct training schemes into a singular, cohesive model. Initially, as the DcAE is designed with the objective of learning encoder-decoder mappings, it endeavors to minimize the squared error between the reconstructed speech and the original input speech. Subsequently, within the code layer, frame-based phonetic embeddings are procured by minimizing the categorical cross-entropy between the ground truth labels and the predicted triphone-state scores. The development of DcAE is grounded in the Kaldi toolkit, wherein various Time Delay Neural Network (TDNN) models are treated as encoders.

In the Chain-based DcAE, they further introduce three novel iterations of the DcAE. Firstly, a new objective function is employed, which takes into consideration both the categorical cross-entropy and the mutual information between ground truth and predicted triphone-state sequences, resulting in the formulation of a chain-based DcAE (c-DcAE). To facilitate its application to robust speech recognition, they further extend c-DcAE to incorporate hierarchical and parallel structures, culminating in the development of hc-DcAE and pc-DcAE, respectively. Within these two models, both the error between the reconstructed noisy speech and the input noisy speech, as well as the error between the enhanced speech and the reference clean speech, are integrated into the objective function.

Experimental results, derived from the Wall Street Journal (WSJ) and Aurora-4 corpora, substantiate that the DcAE models exhibit superior performance when juxtaposed with baseline systems, thereby affirming their efficacy in speech recognition tasks.

\bibliography{rocling2023}
\bibliographystyle{acl_natbib}

\end{document}